\documentclass[10pt,journal,cspaper,compsoc]{IEEEtran}
\usepackage[breaklinks=true,colorlinks,citecolor=blue,linkcolor=blue,urlcolor=blue]{hyperref}
\usepackage{url}            
\usepackage{booktabs}       
\usepackage{amsfonts}       
\usepackage{nicefrac}       
\usepackage{times}
\usepackage{graphicx}
\usepackage{amsmath}
\usepackage{amssymb}
\usepackage{color}
\usepackage{xcolor}
\usepackage{multirow}
\usepackage{overpic}
\usepackage{amssymb}
\usepackage{pifont}
\usepackage{makecell}

\usepackage[nocompress]{cite} 
\usepackage{ragged2e}
\usepackage{cancel}  
\newcommand{\cmark}{\ding{51}}
\newcommand{\xmark}{\ding{55}}
\usepackage{geometry}
\usepackage{subfigure}

\usepackage{silence}
\def\eg{\emph{e.g.,~}}
\def\ie{\emph{i.e.,~}}
\def\etal{{\em et al.}}
\def\etc{\emph{etc}}
\renewcommand{\eqref}[1]{Equ.~(\ref{#1})}
\newcommand{\thRows}[1]{\multirow{3}*{#1}}
\newcommand{\thCols}[1]{\multicolumn{3}{c}{#1}}
\newcommand{\tRows}[1]{\multirow{2}*{#1}}
\newcommand{\tCols}[1]{\multicolumn{2}{c}{#1}}
\newcommand{\foCols}[1]{\multicolumn{4}{c}{#1}}
\newcommand{\myPara}[1]{\vspace{.15in} \noindent\textbf{#1}}
\newcommand{\figref}[1]{Fig.~\ref{#1}}
\definecolor{purple}{rgb}{0.4, 0.0, 0.6}
\newcommand{\tabref}[1]{Tab.~\ref{#1}}
\newcommand{\secref}[1]{Section~\ref{#1}}
\usepackage{bm}

\def\StopG{\cancel{\mathbf{G}}}
\def\ours{SERE}
\newcommand{\tb}[1]{\textbf{#1}}

\def\iBOT{iBOT~\cite{zhou2021ibot}}
\def\DINO{DINO~\cite{caron2021emerging}}
\def\BEiT{BEiT~\cite{bao2021beit}}
\def\MAE{MAE~\cite{he2021masked}}
\def\MAEd{MAE$^\ddag$~\cite{he2021masked}}

\begin{document}

\title{\ours: Exploring Feature Self-relation for Self-supervised Transformer}

\author{Zhong-Yu Li, Shanghua Gao, Ming-Ming Cheng
  \IEEEcompsocitemizethanks{\IEEEcompsocthanksitem 
  The authors are with TMCC, CS, Nankai University, Tianjin 300350, China.
  S. Gao is the corresponding author (shgao@mail.nankai.edu.cn). 
}}

\IEEEtitleabstractindextext{
	\begin{abstract} \justifying
	Learning representations with self-supervision
	for convolutional networks (CNN) has been validated to be effective for vision tasks.
	As an alternative to CNN,
	vision transformers~(ViT) have strong representation ability with 
	spatial self-attention and channel-level feedforward networks.
	Recent works reveal that self-supervised learning helps unleash the great potential of ViT.
	Still, most works follow self-supervised strategies designed for CNN,
	\eg instance-level discrimination of samples,
	but they ignore the properties of ViT.
	We observe that relational modeling on spatial and channel dimensions
	distinguishes ViT from other networks.
	To enforce this property,
	we explore the feature \textbf{SE}lf-\textbf{RE}lation~(\ours) 
	for training self-supervised ViT.
	Specifically, instead of conducting self-supervised learning
	solely on feature embeddings from multiple views,
	we utilize the feature self-relations,
	\ie spatial/channel self-relations, for self-supervised learning.
	Self-relation based learning further enhances the
	relation modeling ability of ViT,
	resulting in stronger representations that
	stably improve performance on multiple downstream tasks.
	Our source code is publicly available at: \href{https://github.com/MCG-NKU/SERE}{https://github.com/MCG-NKU/SERE}.
\end{abstract}
}

\maketitle
\IEEEdisplaynontitleabstractindextext
\IEEEpeerreviewmaketitle

\IEEEraisesectionheading{\section{Introduction}\label{sec:introduction}}

\IEEEPARstart{S}{upervised} training of neural networks thrives on 
many vision tasks at the cost of collecting expensive human-annotations
\cite{russakovsky2015imagenet,lin2015microsoft,Zhou_2017_CVPR}.
Learning visual representations from un-labeled images
\cite{caron2020unsupervised,Chen_2021_CVPR,byol,He_2020_CVPR,SCIENCECHINA-s11432-021-3535-2}
has proven to be an effective alternative to supervised training,
\eg convolutional networks~(CNN) trained with self-supervision have shown
comparable or even better performance than its supervised counterparts
\cite{he2016deep,gao2022rfnext}.
Recently, vision transformers~(ViT)~\cite{dosovitskiy2020vit,liu2021Swin} 
have emerged with stronger representation ability than CNN on many vision tasks.
Pioneering works have shifted the methods designed 
for self-supervised CNN to ViT and revealed the great potential of 
self-supervised ViT~\cite{caron2021emerging,he2021masked,Chen_2021_ICCV}.
Typical self-supervised learning methods designed for ViT, 
\eg \DINO and MoCoV3~\cite{Chen_2021_ICCV},
send multiple views of an image into a ViT network
to generate feature representations.
Self-supervisions, \eg contrastive learning~\cite{Chen_2021_ICCV,xie2021moby,MIR-2022-05-167} 
and clustering~\cite{caron2021emerging,zhou2021ibot},
are then implemented on these representations based on the hypothesis 
that different views of an image share similar representations.
However, the widely used feature representations are still limited to
feature embedding used by CNN based methods, 
\eg image-level embeddings~\cite{chen2020simple,He_2020_CVPR,byol} 
and patch-level embeddings~\cite{wang2020DenseCL,Henaff_2021_ICCV}.
But the properties of ViT, \eg the self-relation modeling ability,
are less considered by existing self-supervised methods.
We wonder if other forms of representations
related to ViT can benefit the training of self-supervised ViT.

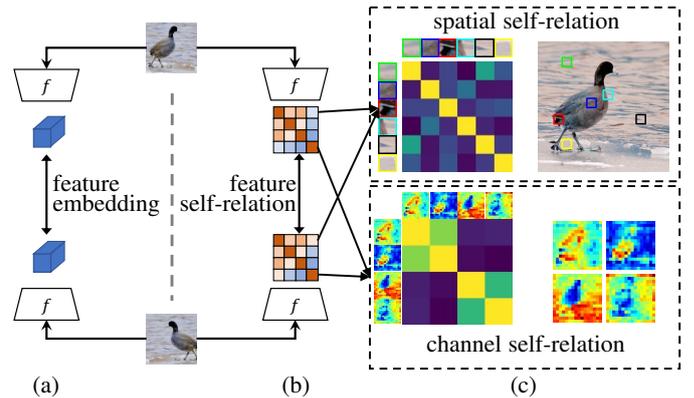
\begin{figure}[t]
	\centering
	\small
	\begin{overpic}[width=1.0\linewidth]{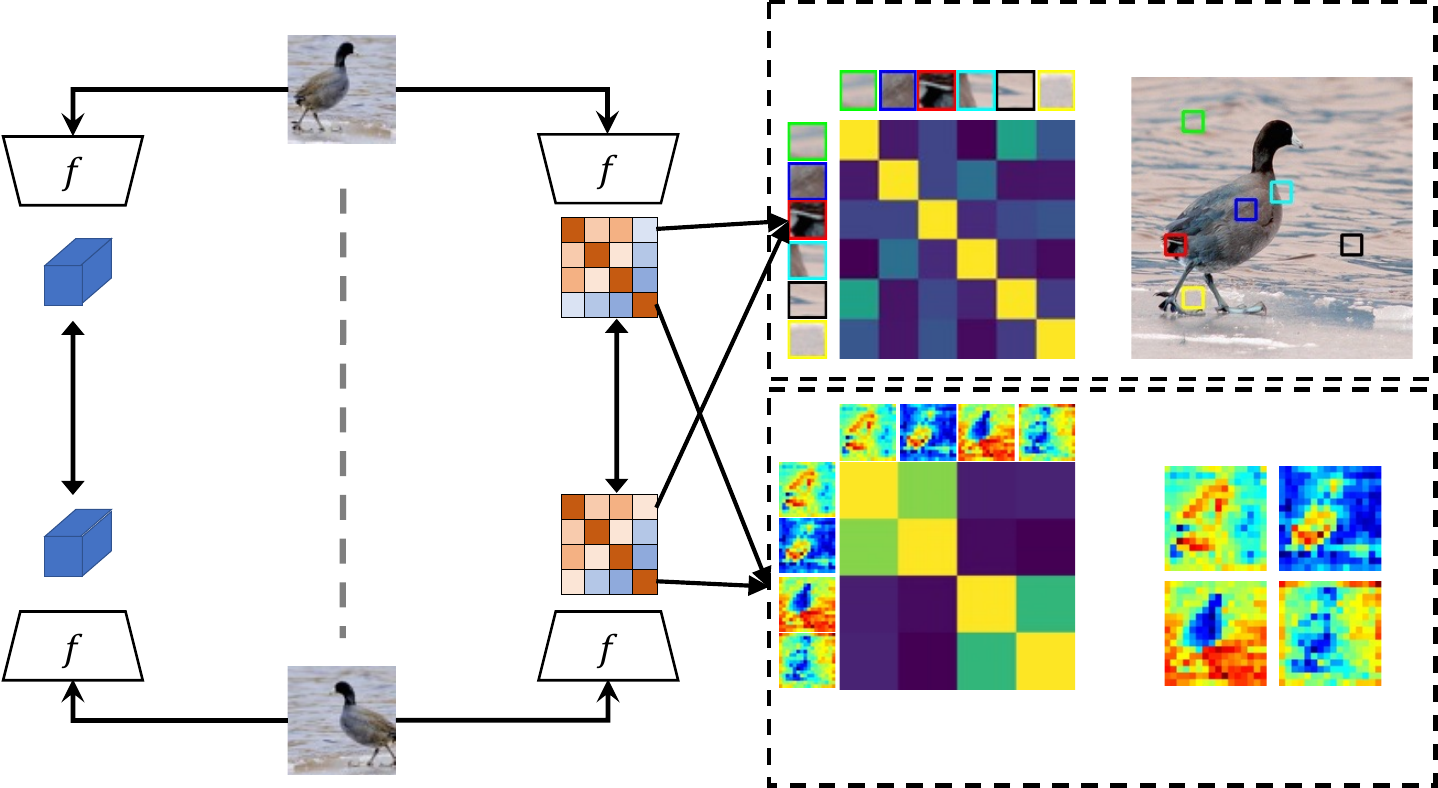}
		\put(3.0, -3.5){(a)}
		\put(40.3, -3.5){(b)}
		\put(74.5, -3.5){(c)}
		\put(6, 27){feature}
		\put(6, 24){embedding}
		\put(32.5, 27){feature}
		\put(25, 24){self-relation}
		\put(62, 2.7){channel self-relation}
		\put(63, 51.5){spatial self-relation}
	\end{overpic}
	\caption{The illustration of self-supervised learning using
		feature embeddings and our proposed feature self-relation.
		(a) Typical self-supervised learning methods process the feature embeddings
		of the image views.
		(b) We propose to model the feature self-relation that measures 
		the relation inside an image view from different dimensions.
		(c) Two specific forms of self-relation, 
		\ie the spatial and channel self-relations.
    For spatial self-relation, we select 6 patches indicated by 
		differently colored boxes~(top right) and 
		visualize their self-relation~(top left).
		For channel self-relation, we show visualized feature maps of 4 channels~(bottom right) 
		and the corresponding self-relation~(bottom left).
	}\label{fig:vis_analysis}
\end{figure}

We seek to improve the training of self-supervised ViT by exploring 
the properties of ViT.
ViT models the feature relations on spatial and channel dimensions 
with the multi-head self-attention (MHSA) and feedforward network
(FFN)~\cite{raghu2021do,dosovitskiy2020vit,Kim_2021_CVPR}, respectively.
The MHSA aggregates the spatial information with the extracted relations 
among patches,
resulting in stronger spatial relations among patches 
with similar semantic contexts (see \figref{fig:vis_analysis}(c)).
The FFN combines features from different channels,
implicitly modeling the feature self-relation in the channel dimension.
For instance, \figref{fig:vis_analysis}(c) reveals that
channels learn diverse patterns, 
and there are varying degrees of relations between different channels. 
Feature self-relation modeling enables ViT with strong representation ability, 
motivating us to use self-relation as a new representation form
for self-supervision.

In this work, we propose to utilize the feature \textbf{SE}lf-\textbf{RE}lation
(\ours)~for self-supervised training,
enhancing the self-relation modeling properties in ViT.
Following the spatial relation in MHSA and channel relation in FFN,
we form the spatial and channel self-relations as representations.
The spatial self-relation extracts the relations among patches within an image.
The channel self-relation models the connection of different channels,
where each channel in feature embeddings highlights unique semantic information.
Feature self-relation is the representation in a new dimension and 
is compatible with existing representation forms, 
\eg image-level and patch-level feature embeddings.
As shown in~\figref{fig:vis_analysis},
we can easily replace the feature embeddings with the proposed feature 
self-relation on existing self-supervised learning methods.
We demonstrate that utilizing feature self-relation could stably
improve multiple training methods for self-supervised ViT, 
\eg \DINO, \iBOT, 
and MoCoV3~\cite{Chen_2021_ICCV},
on various downstream tasks, 
\eg object detection~\cite{Cai_2018_CVPR,lin2015microsoft},
semantic segmentation~\cite{Everingham2009ThePV,Zhou_2017_CVPR}, 
semi-supervised semantic segmentation~\cite{gao2021luss} 
and image classification~\cite{russakovsky2015imagenet}.
To our best knowledge, 
we are the first to study self-relations in self-supervised learning.
Our major contributions are summarized as follows:

\begin{itemize}
	\item We propose to utilize the self-relations (\ours) of ViT, \ie
	  spatial and channel self-relations that fit well with the relation modeling 
		property of ViT, 
	  as the representations for self-supervised learning.
	\item The proposed \ours~method is compatible with 
		existing self-supervised methods and stably boosts ViT 
		on various downstream tasks.
\end{itemize}

\section{Related Work}
\subsection{Self-Supervised Learning}

Self-supervised learning aims at learning rich representations 
without any human annotations.
Early works utilized hand-crafted pretext tasks, 
\eg coloration~\cite{zhang2016colorful,Larsson_2017_CVPR}, 
jigsaw puzzles~\cite{norooziECCV16}, 
rotation prediction~\cite{gidaris2018unsupervised},
autoencoder~\cite{Doersch_2015_ICCV,VincentLBM08}, 
image inpainting~\cite{Pathak_2016_CVPR} and 
counting~\cite{Noroozi_2017_ICCV} 
to learn representations based on heuristic cues~\cite{chen2020simple}, 
but only achieved limited generalization ability. 
Recently, self-supervised learning 
has shown great breakthroughs due to new forms of self-supervisions, 
\eg contrastive learning~\cite{He_2020_CVPR,Wu_2018_CVPR,Zhao_2021_ICCV,
Dwibedi_2021_ICCV,DecoupledContrastive,MIR-2022-08-271,MIR-2022-11-334}, 
self-clustering~\cite{caron2018deep,Zhan_2020_CVPR,YM.2020Self-labelling}, 
and representation alignment~\cite{byol,Chen_2021_CVPR,Koohpayegani_2021_ICCV,
ermolov2021whitening,DynamicsContrastive,ge2021revitalizing}.
These methods directly utilize the feature embeddings as representations 
to generate self-supervisions.
For example, many of these methods utilize 
image-level feature embeddings~\cite{Hu_2021_CVPR,chen2020simple,caron2018deep} 
as representations.
And some methods explore using embeddings in more fine-grained dimensions, 
\eg pixel~\cite{wang2020DenseCL,Xie_2021_CVPR}, 
patch~\cite{Xie_2021_ICCV,Dai_2021_CVPR}, 
object~\cite{Henaff_2021_ICCV}, 
and region~\cite{roh2021scrl,Henaff_2021_ICCV} dimensions.
However, these representations are still embeddings corresponding to 
different regions of input images.
Compared to these embedding based methods that only constrain 
individual embedding,  
we further transform the feature embedding to self-relation 
as a new representation dimension, 
which adds the constraint to the relation among embeddings. 
The self-relation provides rich information for self-supervised training 
and fits well with the relation modeling properties of ViT, 
thus further boosting the representation quality of ViT. 
Meanwhile, the self-relation is orthogonal to embedding based methods and 
consistently improves the performance of multiple methods.

\subsection{Self-Supervised Vision Transformer}
Transformers have been generalized to 
computer vision~\cite{dosovitskiy2020vit,wang2021pyramid} and
achieved state-of-the-art performance on many tasks,
\eg image classification~\cite{liu2021Swin}, 
semantic segmentation~\cite{wang2021pyramid,cheng2021maskformer}, 
and object detection~\cite{wu2022p2t}.
Due to a lack of inductive bias,
training ViT requires much more data and tricks
\cite{dosovitskiy2020vit,touvron2021training}.
Recent works have been working on training ViT with self-supervised learning
methods~\cite{li2022efficient,mugs2022SSL,xie2021moby,li2021mst} 
to meet the data requirement of ViT with low annotation costs.
Many instance discrimination based methods use feature embeddings 
as the representation for self-supervised learning.
For instance, Chen \etal~\cite{Chen_2021_ICCV} and 
Caron \etal~\cite{caron2021emerging} 
implement contrastive learning and self-clustering with image-level embeddings, respectively.
Zhou \etal~\cite{zhou2021ibot} develop self-distillation with patch-level embeddings. 
However, these methods still follow the pretext task of instance discrimination 
initially designed for CNNs, 
where representations with invariance to transformation are learned by
maximizing the similarity among positive samples.
New properties in ViT may help the self-supervised training 
but are ignored by these methods.
We explore spatial self-relation and channel self-relation, 
which are proven more suitable for the training of ViT.

\subsection{Masked Image Modeling}

Concurrent with our work, self-supervised learning by masked image modeling~(MIM)
\cite{Pathak_2016_CVPR,bao2021beit,he2021masked,gao2022towards} 
has become a popular alternative to instance discrimination~(ID) 
for self-supervised ViT. 
MIM reconstructs masked patches from unmasked parts, 
with different forms of reconstruction targets, 
\eg discrete tokenizer~\cite{bao2021beit,chen2022context}, 
raw pixels~\cite{he2021masked,xie2021simmim,repre,li2021mst,atito2021sit}, 
HOG features~\cite{cwei2021}, patch representations~\cite{zhou2021ibot}, \etc. 
Compared to ID, 
patch-level reconstruction in MIM enhances token-level representations~\cite{gao2022towards,zhou2021ibot}. 
Differently, 
the proposed \ours~enhances the ability to model inter-token relations. 
Experiments also demonstrate that \ours~can outperform and complement various MIM-based methods. 
Additionally, we strengthen the ability to model inter-channel relations, 
which MIM is missing.

\subsection{Property of Vision Transformer}

Recent works have shown that the remarkable success of ViT on many vision tasks~\cite{liu2021Swin,carion2020end,cheng2021maskformer}
relies on their strong ability to model spatial relations.
Dosovitskiy \etal~\cite{dosovitskiy2020vit} and 
Kim \etal~\cite{Kim_2021_CVPR} find that attention attends to
semantically relevant regions of images.
Raghu \etal\cite{raghu2021do} reveal the representations of ViT 
preserve strong spatial information even in the deep layer.
They also observe that patches in ViT have strong connections to regions 
with similar semantics.
Caron \etal~\cite{caron2021emerging} find that self-supervised ViT
captures more explicit semantic regions than supervised ViT.
These observations indicate that ViT has a strong ability to model relations,
which is quite different from the pattern-matching mechanisms of CNNs.
In this work, we propose to enhance such ability by explicitly using spatial
and channel feature self-relations for self-supervised learning.

\subsection{Relation Modeling}

Relation modeling, 
which has different forms such as pair-wise relation and attention, 
has facilitated various vision tasks, 
\eg knowledge distillation\cite{Tung_2019_ICCV,Park_2019_CVPR,pkt_eccv,
Peng_2019_ICCV,LocalCorrelation,Zagoruyko2017AT}, 
metric learning~\cite{Chen_Wang_Zhang_2018}, 
semantic segmentation~\cite{Liu_2019_CVPR,He_2019_CVPR,Yang_2022_CVPR}, 
unsupervised semantic segmentation~\cite{hamilton2022unsupervised}, 
object localization~\cite{simeoni2018unsupervised,LOST,Wang_2022_CVPR}, 
contrastive learning~\cite{Ki_2021_ICCV}, 
masked image modeling~\cite{whattohide}, 
feature aggregation~\cite{crow} 
and texture descriptor~\cite{NIPS2015_a5e00132,Lin_2015_ICCV}.
In self-supervised learning, 
early work~\cite{barlow_Twins} proposes to utilize relation modeling 
by calculating channel relations in the whole batch, \ie batch-relation. 
In comparison, we explore self-relation, 
which is the spatial or channel relations for features within an image 
and fits well with the relation modeling property of ViT.

\section{Method}

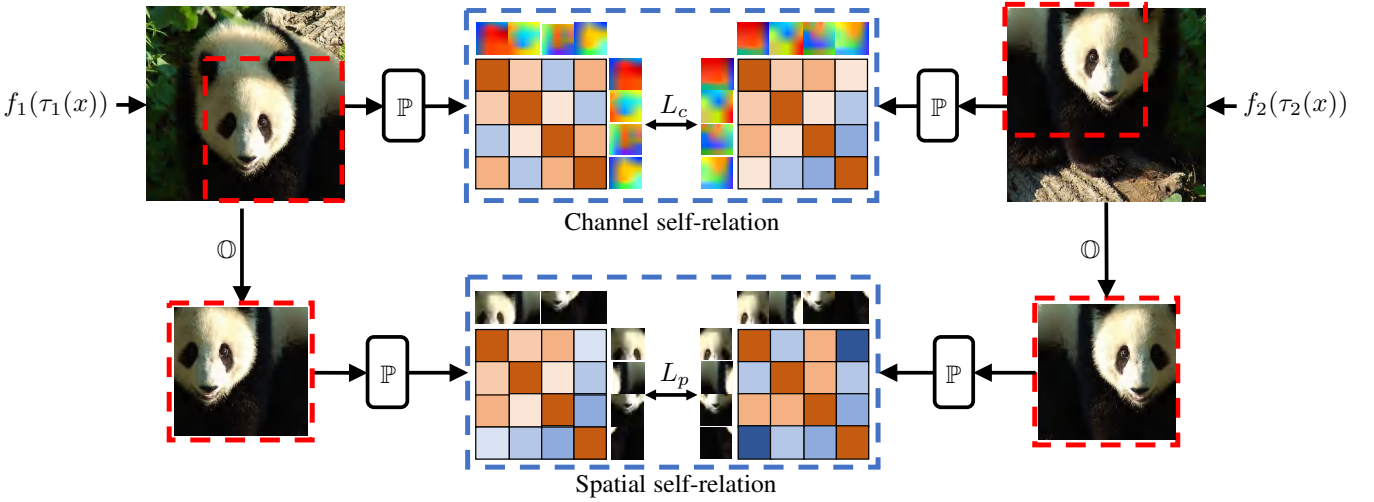
\begin{figure*}[t]
	\centering
	\begin{overpic}[width=.82\linewidth]{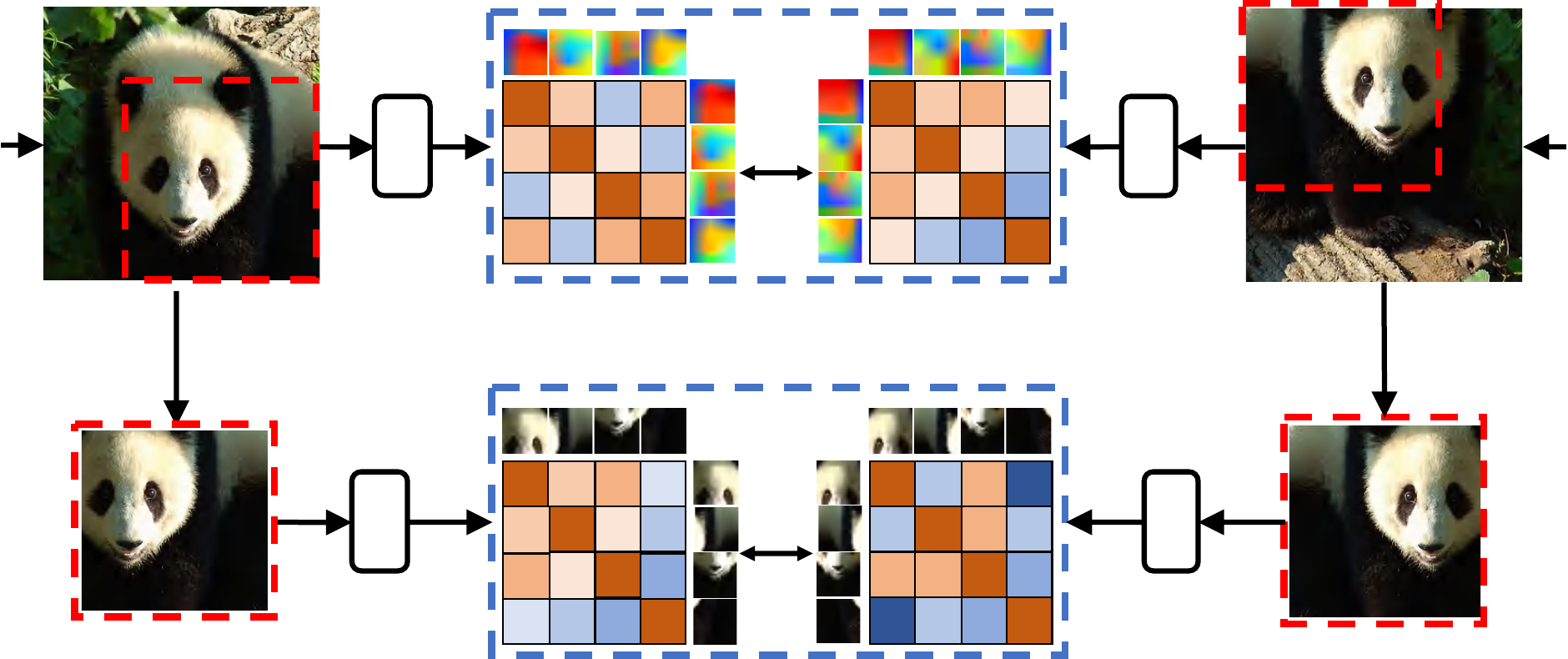}
		\put(40, 21.5){Channel self-relation}
		\put(41, -2){Spatial self-relation}
		
		\put(-10,32.25){$f_1(\tau_1(x))$}
		\put(100.5,32.25){$f_2(\tau_2(x))$}

		\put(23.6, 7.8){$\mathbb{P}$}
		\put(74.0, 7.8){$\mathbb{P}$}
		\put(25.1, 31.8){$\mathbb{P}$}
		\put(72.7, 31.8){$\mathbb{P}$}

		\put(48.5,32){$L_c$}
		\put(48.5,8){$L_p$}

		\put(9,19){$\mathbb{O}$}
		\put(86,19){$\mathbb{O}$}
	\end{overpic}
	\caption{Our method models self-relation from spatial and channel dimensions.
    Given an image $x$, 
		two views are generated by two random data augmentations. 
		Here the image patches represent the feature embeddings extracted by the encoder.
    The feature embeddings are transformed by 
		representation transformation $\mathbb{P}$ to generate spatial or channel self-relations. 
    $L_p$ and $L_c$, \ie the loss functions defined in \eqref{eq:l_b} 
		and \eqref{eq:l_d}, 
  enforce consistency between self-relations of different views.
    For spatial self-relation, 
		only the features in the overlapping region are considered.
		$\mathbb{O}$ means the operation of extracting features from the 
		overlapping region between two views in \eqref{eq:self_relation_pixel}, 
		where the red dotted box indicates the overlapping region.
	}	\label{fig:vis_framework}
\end{figure*}

\subsection{Overview}
\label{sec:overview}

In this work, we focus on the instance discriminative self-supervised 
learning pipeline \cite{caron2020unsupervised,caron2021emerging}.
First, we briefly revisit the framework of common
instance discriminative self-supervised learning methods.
Given an un-labeled image $x$,
multiple views are generated by different random data augmentations,
\eg generating two views $\tau_{1}(x)$ and $\tau_{2}(x)$ 
with augmentations $\tau_{1}$ and $\tau_{2}$.
Under the assumption that different views of an image contain 
similar information,
the major idea of most instance discriminative methods is to maximize the
shared information encoded from different views.
Firstly,
two views are sent to the encoder network to extract the feature embeddings 
$r_1 \in \mathbb{R}^{C\times HW}$ and $r_2 \in \mathbb{R}^{C\times HW}$
with $H \cdot W$ local patches and $C$ channels.
According to the training objective of self-supervised learning methods,
the feature embeddings are then transformed with transformation 
$\mathbb{P}$ to obtain different representations, 
\eg image-level and patch-level embeddings.
Different self-supervised optimization objectives utilize
the obtained representations to get the loss as follows:
\begin{equation}
	L_I= {\rm R}(\mathbb{P}(r_1), \mathbb{P}(r_2)),
	\label{eq:l_I}
\end{equation}
where ${\rm R}$ means the function that maximizes the consistency across views 
and can be defined with multiple forms,
\eg contrastive~\cite{He_2020_CVPR}, non-contrastive~\cite{byol}, 
and clustering~\cite{caron2020unsupervised} losses.

Our main focus in this work is exploring new forms of
representation transformation $\mathbb{P}$.
Motivated by the relation modeling properties in ViT,
instead of directly using feature embeddings,
we utilize feature self-relation in multiple dimensions
as the representations for self-supervised learning on ViT.
In the following sections, we introduce
two specific self-relation representations
for self-supervised ViT, \ie spatial and channel self-relations.

\begin{figure}
    \centering
    \begin{overpic}[width=.85\linewidth]{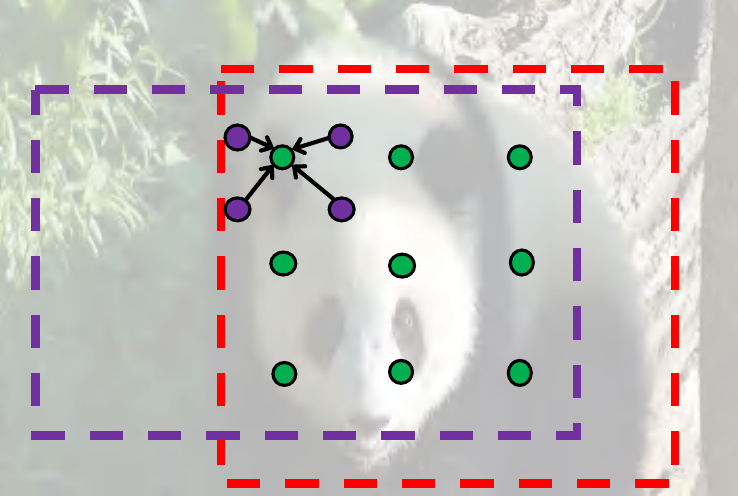}
    \put(8, 60){\color{purple} \bm{$\tau_1(x)$}}
    \put(34, 61.5){\color{red} \bm{$\tau_2(x)$}}
    \end{overpic}
    \caption{The region-aligned sampling operation for spatial self-relation. 
    $\tau_1(x)$ and $\tau_2(x)$ are the different views of an image, 
    and the dotted boxes indicate their regions in the original image. 
    The {\color{green}points} in green mean the uniformly sampled points in the overlapped regions.
    And the {\color{purple}points} in purple mean the patch features in ViT.} 
    \label{fig:sampling}
\end{figure}

\subsection{Spatial Self-relation}
\label{sec:pixel_relation}

Prior works~\cite{dosovitskiy2020vit,Kim_2021_CVPR,raghu2021do,caron2021emerging} 
have observed that ViT has the property of modeling relations 
among local patches by the MHSA module.
Meanwhile, modeling more accurate spatial relations 
is crucial for many dense prediction tasks
\cite{wang2020DenseCL,Henaff_2021_ICCV},
\eg object detection and semantic segmentation.
So we propose to enhance the relation modeling ability of ViT
by cooperating spatial self-relation for self-supervised training.
In the following part,
we first give details of the transformation $\mathbb{P}$ that transforms the feature embeddings encoded by ViT 
to spatial self-relation.
Then, we give the self-supervision loss
utilizing spatial self-relation as the representation.

\myPara{Generating spatial self-relation representation.}
\label{par:pixel}
Given the feature embeddings $r_1 = f_1(\tau_1(x)) \in \mathbb{R}^{C\times HW}$ 
and $r_2 = f_2(\tau_2(x)) \in \mathbb{R}^{C\times HW}$
from the ViT backbone,
a projection head $h_p$,
which consists of a batch normalization~\cite{batchnorm} layer and 
a ReLU~\cite{agarap2018deep} activation layer, 
processes these
embeddings to obtain $p_{1} = h_p(r_1)$ and $p_{2} = h_p(r_2)$.
Then, we separately calculate their spatial self-relation.

In contrast to the image-level embedding, 
the supervision between spatial self-relation of different views 
should be calculated between patches at the same spatial positions. 
However, 
$p_1$ and $p_2$ are not aligned 
in the spatial dimension due to the random crop and flip in data augmentations.
To solve the misalignment issue, 
we apply a region-aligned sampling operation $\mathbb{O}$~\cite{gao2021luss}
to uniformly sample $H_s\times W_s$ points 
from the overlapping region of $p_1$ and $p_2$.\footnote{
In this work, we combine the proposed spatial self-relation 
with existing methods due to the orthogonality of self-relation. 
Since existing methods do not restrict 
that different views must overlap, 
we only add spatial self-relation to the views with overlapping regions.} 
As shown in \figref{fig:sampling}, 
we localize the overlapping region in the raw image and split the region into $H_s\times W_s$  grids, 
which are not essentially aligned with the patches in ViT. 
For the center of each grid, we calculate its spatial coordinates in feature maps of each view and 
then sample its features by bi-linear interpolation.
The details of this operation $\mathbb{O}$ are shown in the supplementary.
For one view,
\eg $p_1\in \mathbb{R}^{C\times HW}$,
we calculate the spatial self-relation $\mathbb{A}_{p}(p_1) \in \mathbb{R}^{H_sW_s \times H_sW_s}$  as follows: 
\begin{align}
	\mathbb{A}_{p}(p_1) = {\rm Softmax} \left( \frac{\mathbb{O}(p_1)^{T}\cdot \mathbb{O}(p_1)}{\sqrt{C}}\bigg/t_p \right ),
	\label{eq:self_relation_pixel}
\end{align}
where $\mathbb{O}(p_1)\in R^{C\times H_sW_s}$ is the feature 
sampled in the overlapping region,
$T$ is the matrix transpose operation,
and $t_p$ is the temperature parameter that controls
the sharpness of the ${\rm Softmax}$ function.
In the spatial self-relation, each row represents the relation of one local patch to other patches
and is normalized by the ${\rm Softmax}$ function to generate probability distributions.

\myPara{Self-supervision with spatial self-relation.}
Spatial self-relation can be used
as the representation of many
forms of self-supervisions.
For simplicity, we give an example of using self-relation
for asymmetric non-contrastive self-supervision loss~\cite{byol,Chen_2021_CVPR}
as follows:
\begin{equation}
	L_{p} = {\rm R}_{e}(\StopG(\mathbb{A}_p(p_1)), \mathbb{A}_p(g_p(p_2))),
	\label{eq:l_b}
\end{equation}
where ${\rm R}_{e}$ is the cross-entropy loss, 
$\StopG$ is the stop-gradient operation
to avoid training collapse following~\cite{Chen_2021_CVPR},
and $g_p$ is the prediction head 
for asymmetric non-contrastive loss~\cite{byol,Chen_2021_CVPR} consisting of
a fully connected layer,
a batch normalization layer, and a ReLU layer.

\myPara{Multi-head spatial self-relation.}
\label{par:multihead}
In ViT, the MHSA performs multiple parallel self-attention operations 
by dividing the feature into multiple groups.
It is observed that different heads might focus on 
different semantic patterns~\cite{caron2021emerging}.
Inspired by this,
we divide the feature embeddings into $M$ groups along 
the channel dimension and 
calculate the spatial self-relation within each group,
obtaining $M$ spatial self-relations for each view.  
By default, we choose $M=6$, as shown in \tabref{tab:multi_head}.

\subsection{Channel Self-relation}

In neural networks, 
each channel represents some kind of pattern within images. 
Different channels encode diverse patterns~\cite{Liu_2021_ICCV,NIPS2012_c399862d}, 
providing neural networks with a strong representation capability.
The FFN~\cite{dosovitskiy2020vit} in ViT combines patterns across channels and
implicitly models the relation among channels~\cite{Liu_2021_ICCV},
\ie the pattern encoded in one channel has
different degrees of correlation with the patterns encoded by other channels,
as shown in~\figref{fig:vis_framework}.
This mechanism motivates us to form channel self-relation as 
the representation for self-supervised learning to 
enhance self-relation modeling ability in the channel dimension.
Specifically, we transform the feature embedding of ViT to channel self-relation
and then use the channel self-relation
as the representation for self-supervision.

\myPara{Generating channel self-relation representation.}
Here, we give the details of the transformation $\mathbb{P}$ 
that transforms the feature embeddings to channel self-relation.
As in~\eqref{eq:self_relation_pixel},
given the feature embeddings of two views, \ie $r_1$ and $r_2$,
a projection head $h_c$
with the same structure as $h_p$ 
processes these embeddings and obtains $c_1=h_c(r_1)^T$ and $c_2=h_c(r_2)^T$.
Then we separately calculate the channel self-relation for each view.
For one view, \eg $c_1\in \mathbb{R}^{HW\times C}$,
we calculate its channel self-relation 
$\mathbb{A}_{c}(c_1) \in \mathbb{R}^{C\times C}$ as follows:
\begin{equation}
	\mathbb{A}_{c}(c_1)={\rm Softmax} \left( \frac{c_1^{T} \cdot c_1}{H \cdot W}\bigg/t_c \right ),
	\label{eq:self_relation_channel}
\end{equation}
where the Softmax function normalizes each row of the self-relation to get probability distributions,
and $t_c$ is the temperature parameter 
controlling the sharpness of
probability distributions.

\myPara{Self-supervision with channel self-relation.}
The channel self-relation
can also be utilized as a new form of representation
for many self-supervised losses.
Similar to the spatial self-relation based loss in~\eqref{eq:l_b}, we give
the non-contrastive loss using channel self-relation as follows:
\begin{equation}
	L_{c} = {\rm R_{e}}(\StopG(\mathbb{A}_{c}(c_1)), \mathbb{A}_{c}(g_c(c_2))),
	\label{eq:l_d}
\end{equation}
where the $ {\rm R_{e}}$ is the cross-entropy loss,
and $g_c$ is a prediction head with the same structure as $g_p$
in~\eqref{eq:l_b}.
This loss function enforces the consistency of channel self-relations among views
and thus enhances the channel self-relation
modeling ability of the model.
Unlike spatial self-relation, 
we do not need to consider the spatial misalignment between different views. 
Because we enforce the consistency between channel self-relations, 
not the channel features, 
and the channel self-relation defined in \eqref{eq:self_relation_channel} has no spatial dimension.

\subsection{Implementation Details}
\label{sec:imp_details}
\myPara{Loss function.}
By default, we apply our proposed spatial/channel self-relations and image embeddings as representations for self-supervision losses,
as these representations reveal different properties of features.
The summarized loss function is as follows:
\begin{equation}
	L = L_I + \alpha L_p + \beta L_c,
	\label{eq:loss_sum}
\end{equation}
where the spatial and channel losses are weighted by $\alpha$ and $\beta$, 
and $L_I$ is the loss using image-level embeddings, \eg the clustering-based loss in \DINO.
We show in~\tabref{tab:ablation_channel_pixel} 
that solely using our proposed self-relation could achieve competitive or better performance than using image-level embeddings.
Combining these three representations results in better representation quality,
showing self-relation is a complementary representation form to image-level embeddings.
To increase the training efficiency
and make fair comparisons, 
we utilize the multi-crop~\cite{caron2021emerging,caron2020unsupervised} augmentation to generate global and local views.
For local views,
we follow~\cite{caron2021emerging,caron2020unsupervised}
to calculate the loss
between each global and local view 
but ignore the loss among local views.

\myPara{Architecture.}
We use the Vision Transformer~\cite{dosovitskiy2020vit}
as the encoder network.
Following~\cite{He_2020_CVPR,caron2021emerging},
the representations $r_1$ and $r_2$ of two views $\tau_1(x)$ and $\tau_2(x)$ 
are extracted
by a momentum-updated encoder network $f_1$ and the encoder network $f_2$. 
During training, the parameters $\theta_2$ of $f_2$ are updated by gradient descent.
And the parameters $\theta_1$ of $f_1$ are updated as $\theta_{1} = \lambda \theta_{1} + (1 - \lambda) \theta_{2}$,
where $\lambda \in [0, 1]$ is the momentum coefficient.
Following \DINO,
the $\lambda$ is set to 0.996 and is increased to 1.0 during training with a cosine schedule.
Accordingly, we denote the projections following $f_1$ and $f_2$ as $h_p^1$/$h_c^1$ and $h_p^2$/$h_c^2$, respectively. 
The parameters of $h_p^1$/$h_c^1$ are also momentum-updated by $h_p^2$/$h_c^2$, following the updating scheme of $f_1$.
Only the encoder network is used for transfer learning
on downstream tasks after pre-training.

\section{Experiments}
This section verifies the effect of 
using proposed spatial 
and channel self-relations as representations for self-supervised learning.
We give the pre-training settings in~\secref{sec:settings}.
In~\secref{sec:performance}, we compare our method with existing methods on multiple evaluation protocols,
showing stable improvement over multiple methods.
In~\secref{sec:Ablation}, we conduct ablations to clarify design choices.

\begin{table}[t]
	\caption{Fully fine-tuning classification on ImageNet-1K and semi-supervised semantic segmentation on ImageNet-S.	
        For ImageNet-S, we report the mIoU on the val and test set.
        The PT means loading self-supervised pre-trained weights for initialization 
        and FT means loading fully fine-tuned weights on classification labels of ImageNet-1K for initialization, respectively.}
	\centering
	\setlength{\tabcolsep}{0.4mm}
	\begin{tabular}{lcccc|cccc}	\toprule
		& \thRows{Backbone} & \thRows{Epochs} & \tCols{Classification} \vline & \foCols{Segmentation} \\ 
		& & & \tCols{ImageNet-1K} \vline
		& \tCols{ImageNet-S$_{\rm PT}$} & \tCols{ImageNet-S$_{\rm FT}$} \\ 
		& & & Top-1 & Top-5 & val               & test         & val          & test             \\
		\Xhline{0.8pt}
		\DINO & ViT-S/16 & 100 & 79.7 & 95.1 & 35.1          & 34.4   &  54.6 & 54.4    \\
		+\ours                        & ViT-S/16 & 100 & \textbf{80.9} & \textbf{95.5} & \textbf{36.9} & \textbf{36.0} & \textbf{57.3} & \textbf{56.2}        \\ 
		\hline
		\iBOT & ViT-S/16& 100  & 80.9 & 95.4 & 38.1 & 37.8 & 57.9 & 57.4 \\
		+\ours                   & ViT-S/16& 100  & \textbf{81.5} & \textbf{95.8} & \textbf{41.0} & \textbf{40.2}  & \textbf{58.9} &  \textbf{57.8} \\
        \hline
		\iBOT & ViT-B/16 & 100 & 83.3 & 96.6 & 48.3 & 47.8 & 62.6 & 63.0 \\
		+\ours                   & ViT-B/16 & 100 & \tb{83.7} & \tb{96.7} & \tb{48.6} & \tb{48.2} & \tb{63.0} & \tb{63.3} \\
		\bottomrule
	\end{tabular}
	\label{tab:com_imagenets}
\end{table}

\begin{table}[t]
	\caption{
	Transferring learning on semantic segmentation, object detection, and instance segmentation. 
        The AP$^{\rm b}$ means the bounding box AP for object detection~(DET), and AP$^{\rm m}$ means the segmentation mask AP for instance segmentation~(SEG).}
	\centering
        \setlength{\tabcolsep}{2.5mm}
        \begin{tabular}{lccccccccccc}	\toprule
                                     & \thCols{VOC SEG}   & \thCols{ADE20K SEG} \\ \cline{2-7}
                                     & mIoU       &       & mAcc               & mIoU     &     & mAcc  \\ 
            \midrule
        \DINO & 77.1         &     & 87.5               & 42.6     &      & 53.4          \\
        +\ours                         & \textbf{79.7}  &    & \textbf{88.8}      & \textbf{43.8} & & \textbf{54.6}   \\
            \midrule
            & \thCols{COCO DET} & \thCols{COCO SEG} \\ \cline{2-7}
            & AP$^{\rm b}$  & AP$^{\rm b}_{\rm 50}$ & AP$^{\rm b}_{\rm 75}$ & AP$^{\rm m}$  & AP$^{\rm m}_{\rm 50}$ & AP$^{\rm m}_{\rm 75}$ \\ \midrule
        \DINO &  46.0          & 64.9                  & 49.7                  & 40.0          & 62.0                  & 42.8                  \\
        +\ours                         & \textbf{46.6} & \textbf{65.9}         & \textbf{50.2}         & \textbf{40.5} & \textbf{62.9}         & \textbf{43.5}         \\
            \bottomrule
        \end{tabular}
        \label{tab:1k_downstream}
\end{table}

\begin{table}[t]
    \caption{Comparison with longer pre-training epochs.}
    \label{tab:com_more_epochs}
    \centering
    \vspace{-10pt}
    \subtable[Semantic segmentation on the ADE20K dataset.]{
    	\setlength{\tabcolsep}{3.0mm}
    	\begin{tabular}{lcccc}	\toprule
    		& Backbone & Epochs &  mIoU & mAcc \\ 
    		\hline
    		\iBOT & ViT-S/16 & 800  & 45.4 & 56.2 \\
    		+\ours  & ViT-S/16 & 100  & \tb{45.8} & \tb{56.8} \\ \hline
    		\iBOT & ViT-B/16 & 400  & 50.0 & 60.3 \\
    		+\ours                   & ViT-B/16 & 200  & \tb{50.0} & \tb{60.9} \\
    		\bottomrule
    	\end{tabular}
	\label{tab:com_ade_more_epochs}
    }
    \subtable[Classification on the ImageNet-1K dataset.]{
    	\setlength{\tabcolsep}{3.0mm}
    	\begin{tabular}{lcccc}	\toprule
    		& Backbone & Epochs &  Top-1 & Top-5 \\ 
    		\Xhline{0.8pt}
    		\iBOT & ViT-S/16 & 300  & 81.1 & - \\
    		+\ours                   & ViT-S/16 & 100  & \tb{81.5} & \tb{95.8} \\
    		\bottomrule
    	\end{tabular}
    	\label{tab:com_in1k_more_epochs}
    }
\end{table}

\subsection{Pre-training Settings}
\label{sec:settings}
Unless otherwise stated,
we adopt the ViT-S/16 as the backbone network.
\DINO is selected as our major baseline method.
The model is trained by an AdamW~\cite{loshchilov2018decoupled} optimizer
with a learning rate of 0.001 and a batch size of 512.
We pre-train models for 100 epochs on the ImageNet-1K~\cite{russakovsky2015imagenet} dataset for performance comparison.
For ablation, the ImageNet-S$_{300}$ dataset~\cite{gao2021luss}
is used to save training costs.
Following~\cite{caron2021emerging},
we apply the multi-crop training scheme where 2 global views with the resolution of 224$\times$224 and 
4 local views with the resolution of
96$\times$96 are adopted.
The global views are cropped with a ratio between 0.35 and 1.0.
And the local views are cropped with a ratio between 0.05 and 0.35.
For spatial self-relation, the $H_s$/$W_s$ of the operation
$\mathbb{O}$ in \eqref{eq:self_relation_pixel}
are set to 13/13 for global views and 6/6 for local views.
The number of heads $M$ in spatial self-relation is set to 6 by default.
The $t_p$ in~\eqref{eq:self_relation_pixel} and $t_c$ in
~\eqref{eq:self_relation_channel}
are set to 0.5 and 0.1 for the encoder network.
For the momentum encoder, 
we set the $t_p$ and $t_c$ to 1.0 and 1.0.
The $\alpha$ and $\beta$ in~\eqref{eq:loss_sum} 
are set to 1.0 and 1.0, respectively.

For \iBOT, 10 local views are used for a fair comparison.
And we crop images with a ratio between 0.4 and 1.0 for global views 
and between 0.05 and 0.4 for local views.
A gradient clip of 0.3 is used for optimization.
The $\alpha$ and $\beta$ in~\eqref{eq:loss_sum} are set to 0.2 and 0.5. 
Additionally, we provide experiments with ViT-B/16 as the backbone 
and show the pre-training and fine-tuning details in the supplementary.

\subsection{Performance and Analysis}
We verify the effectiveness of self-relation
for self-supervised learning
by transferring the pre-trained models
to image-level classification tasks
and dense prediction downstream tasks.
Models are pre-trained with 100 epochs
on ImageNet-1k unless otherwise stated.
For easy understanding,
models pre-trained
with self-relation representations
are marked as SERE.
\label{sec:performance}

\myPara{Fully fine-tuning classification on ImageNet-1K.}
We compare the fully fine-tuning
classification performance on the ImageNet-1K dataset.
When utilizing ViT-S/16, 
the pre-trained model is fine-tuned 
for 100 epochs 
with the AdamW~\cite{loshchilov2018decoupled} optimizer and a batch size of 512.
The initial learning rate is set to 1e-3 with a layer-wise decay of 0.65.
After a warmup of 5 epochs, the learning rate gradually decays to 1e-6 with the cosine decay schedule.
We report the Top-1 and Top-5 accuracy for evaluation on the ImageNet-1k val set. 
As shown in \tabref{tab:com_imagenets}, \ours~advances DINO and iBOT by 1.2\% and 0.6\% on Top-1 accuracy. 
Even compared to iBOT of 300 epochs, 
\ours~can improve 0.4\% Top-1 accuracy 
with a third of the pre-training time~(100 epochs), as shown in \tabref{tab:com_more_epochs} (b). 
Moreover, using ViT-B/16, \ours~surpasses iBOT by 0.4\% in Top-1 accuracy, 
as shown in \tabref{tab:com_imagenets}. 
These results demonstrate that 
\ours~enhances the category-related representation ability of ViT.

\begin{table}[t]
	\caption{Semi-supervised classification on ImageNet-1K. 
        We fine-tune the models with 1\%/10\% training labels and evaluate them with 100\% val labels. 
 }
	\centering
	\setlength{\tabcolsep}{3.8mm}
	\begin{tabular}{lccccccccccc}	\toprule
	    &  \tCols{1\%} &  \tCols{10\%} \\ \cline{2-5}
	    & Top-1               & Top-5 & Top-1               & Top-5 \\
		\midrule
		\DINO & 52.1 & 77.8 & 70.0 & 89.8 \\
		+\ours                         & \textbf{55.9} & \textbf{81.0} & \textbf{71.5} & \textbf{90.6} \\
		\bottomrule
	\end{tabular}
	\label{tab:com_imagenet_semi}
\end{table}

\begin{table}[t]
	\caption{Transfer learning on the classification task. We fine-tune the pre-trained models on multiple datasets and report the Top-1 accuracy.}
	\centering
	\setlength{\tabcolsep}{2.8mm}
	\begin{tabular}{lccccccccccc}	\toprule
	    &  Cifar$_{10}$ & Cifar$_{100}$ & INat$_{19}$ & Flwrs & Cars \\
		\midrule
		\DINO & 98.8 & 89.6 & 76.9 & 97.8 & 93.5 \\
		+\ours                         & \textbf{98.9} & \textbf{90.0} & \textbf{77.5} & \textbf{98.0} & \textbf{93.5} \\
		\bottomrule
	\end{tabular}
	\label{tab:com_other}
\end{table}

\newcommand{\Rows}[2]{\multirow{#1}*{#2}}

\begin{table}[t]
	\caption{Compared with masked image modeling on the ImageNet-1K dataset. 
        $^\dag$ means effective pre-training epochs~\cite{zhou2021ibot} 
        that account for actually used images during pre-training. 
        $^\ddag$ means the models are fine-tuned for 200 epochs on ImageNet-1K, 
        while others are fine-tuned for 100 epochs.}
	\centering
	\setlength{\tabcolsep}{2.0mm}
	\begin{tabular}{lccccc}	\toprule
		& {Architecture} & {Pre-training Epochs$^\dag$} & Top-1 \\ \hline
    \DINO & \Rows{5}{ViT-S/16} & 300 & 79.7 \\
		\MAEd &                    & 800 & 80.9 \\
		\iBOT &                    & 400 & 80.9  \\
    \DINO+\ours &              & 300 & 80.9 \\
		\iBOT+\ours &              & 400 & \tb{81.5} \\ \hline
		\BEiT & \Rows{4}{ViT-B/16} & 800 & 83.2 \\
		\MAE  &                    & 800  & 83.3 \\
		\iBOT &                    & 400  & 83.3 \\
		\iBOT+\ours &              & 400  & \tb{83.7} \\ \bottomrule
	\end{tabular}
	\label{tab:mae}
\end{table}

\def\MoCov{MoCov3~\cite{Chen_2021_ICCV}}

\begin{table}[t]
	\caption{Cooperating SERE with multiple self-supervised learning methods. Models are pre-trained on the ImageNet-S$_{300}$ dataset with 100 epochs.}
	\centering
	\setlength{\tabcolsep}{4.0mm}
	\begin{tabular}{lcccc}	\toprule
    & \tCols{VOC SEG} & \tCols{ImageNet-S$_{300}^{\rm PT}$} \\ \cline{2-5}
    & mIoU & mAcc & val & test \\ \midrule
		\MoCov & 65.7 & 78.7 & 24.0 & 24.8 \\
		+\ours & \tb{67.5} & \tb{80.6} & \tb{29.1} & \tb{29.9} \\ \midrule
		\DINO  & 68.1 & 81.1 & 28.8 & 29.6 \\
		+\ours & \tb{73.5} & \tb{84.7}  & \tb{41.2} & \tb{42.0} \\ \midrule
		\iBOT  & 74.5 & 85.5 & 41.5 & 42.0 \\
		+\ours & \tb{75.9} & \tb{86.3} & \tb{45.3} & \tb{45.6} \\ \bottomrule
	\end{tabular}
	\label{tab:mococv3}
\end{table}

\myPara{Semi-supervised classification on ImageNet-1K.}
We also evaluate the classification performance in a semi-supervised fashion.
Following the setting of~\cite{zhou2021ibot}, 
we fully fine-tune the pre-trained models with 1\% and 10\% training labels on the ImageNet-1K dataset for 1000 epochs. %
We use the AdamW optimizer to train the model with a batch size of 1024 and a learning rate of 1e-5.
\tabref{tab:com_imagenet_semi} reports the Top-1 and Top-5 accuracy on the ImageNet-1K val set.
SERE consistently achieves better accuracy with 1\% and 10\% labels.
With only 1\% labels, there is a significant improvement of 3.8\% in Top-1 accuracy, 
showing the advantage of our method in the semi-supervised fashion.

\myPara{Semi-supervised semantic segmentation for ImageNet-S.}
The ImageNet-S dataset~\cite{gao2021luss} extends ImageNet-1K with pixel-level semantic segmentation annotations on almost all val images
and parts of training images.
Evaluating semantic segmentation on the ImageNet-S dataset
avoids the potential influence of domain shift between pre-training
and fine-tuning datasets. 
We fine-tune the models with the semantic segmentation
annotations in the ImageNet-S training set and evaluate the performance on the val and test sets of ImageNet-S.
The ViT-S/16 model is initialized with self-supervised pre-trained weights~(ImageNet-S$_{\rm PT}$) or fully fine-tuned weights on classification
labels~(ImageNet-S$_{\rm FT}$) of the ImageNet-1K dataset.
A randomly initialized $1\times 1$ conv is attached to 
the model
as the segmentation head.
We fine-tune models for 100 epochs with an AdamW optimizer, using a batch size of 256 and a weight decay of 0.05.
The learning rate is initially set to 5e-4 with a layer-wise decay of 0.5.
After a warmup of 5 epochs, 
the learning rate decays to 1e-6 by the cosine decay schedule.
The images are resized and cropped to 224$\times$224 for training 
and are resized to 256 along the smaller side for evaluation.

As shown in \tabref{tab:com_imagenets}, 
compared to DINO and iBOT, 
SERE improves the val mIoU by 1.8\% and 2.9\% 
when initializing the model with self-supervised pre-trained weights. 
When loading weights of the fully fine-tuned classification model for initialization,
SERE brings a 2.7\%/1.0\% gain on mIoU over DINO/iBOT.
We conclude that \ours~enhances the relation modeling ability,
enabling ViT with much stronger shape-related representations.

\myPara{Transferring learning on the classification task.}
To evaluate the transferring ability on classification tasks, 
we fine-tune pre-trained models on multiple datasets, including CIFAR~\cite{Krizhevsky_2009_17719}, Flowers~\cite{Nilsback2008AutomatedFC}, Cars~\cite{cars}, and iNaturalist19~\cite{Horn_2018_CVPR}.
The training details are summarized in the supplementary.
\tabref{tab:com_other} shows that
SERE performs better on Top-1 accuracy over DINO,
demonstrating that \ours~benefits the transferring learning on classification tasks.

\myPara{Transfer learning on semantic segmentation.}
We also evaluate the transfer learning performance on the semantic segmentation
task using PASCAL VOC2012~\cite{Everingham2009ThePV} and ADE20K~\cite{Zhou_2017_CVPR} datasets.
The UperNet~\cite{Xiao_2018_ECCV} with the ViT-S/16 backbone
is used as the segmentation model.
Following the training setting in \cite{zhou2021ibot},
we fine-tune models for 20k and 160k iterations on PASCAL VOC2012 and ADE20K datasets, 
with a batch size of 16.
\tabref{tab:1k_downstream} reports the mIoU and mAcc on the validation set.
The self-relation improves the DINO by 2.6\% on mIoU and 1.3\% on mAcc for the PASCAL VOC2012 dataset.
On the ADE20K dataset, 
there is also an improvement of 1.2\% on mIoU and 1.2\% on mAcc compared to DINO. 
\tabref{tab:com_more_epochs} (a) 
shows that \ours~even outperforms iBOT with much fewer pre-training epochs.
Therefore,
semantic segmentation tasks
benefit from
the stronger self-relation
representation ability of SERE.

\myPara{Transfer learning on object detection and instance segmentation.}
We use the Cascade Mask R-CNN~\cite{Cai_2018_CVPR} with ViT-S/16 to evaluate the transfer learning performance 
on object detection and instance segmentation tasks.
Following \cite{zhou2021ibot},
the models are trained on the COCO train2017 set~\cite{lin2015microsoft} 
with the 1$\times$ schedule and a batch size of 16.
\tabref{tab:1k_downstream} reports the bounding box AP~(AP$^{\rm b}$) and the segmentation mask AP~(AP$^{\rm m}$) on the COCO val2017 set. 
Compared to DINO, \ours~improves by 0.6\% on AP$^{\rm b}$ and 0.5\% on AP$^{\rm m}$, 
showing that \ours~facilitates the model to locate and segment objects accurately.

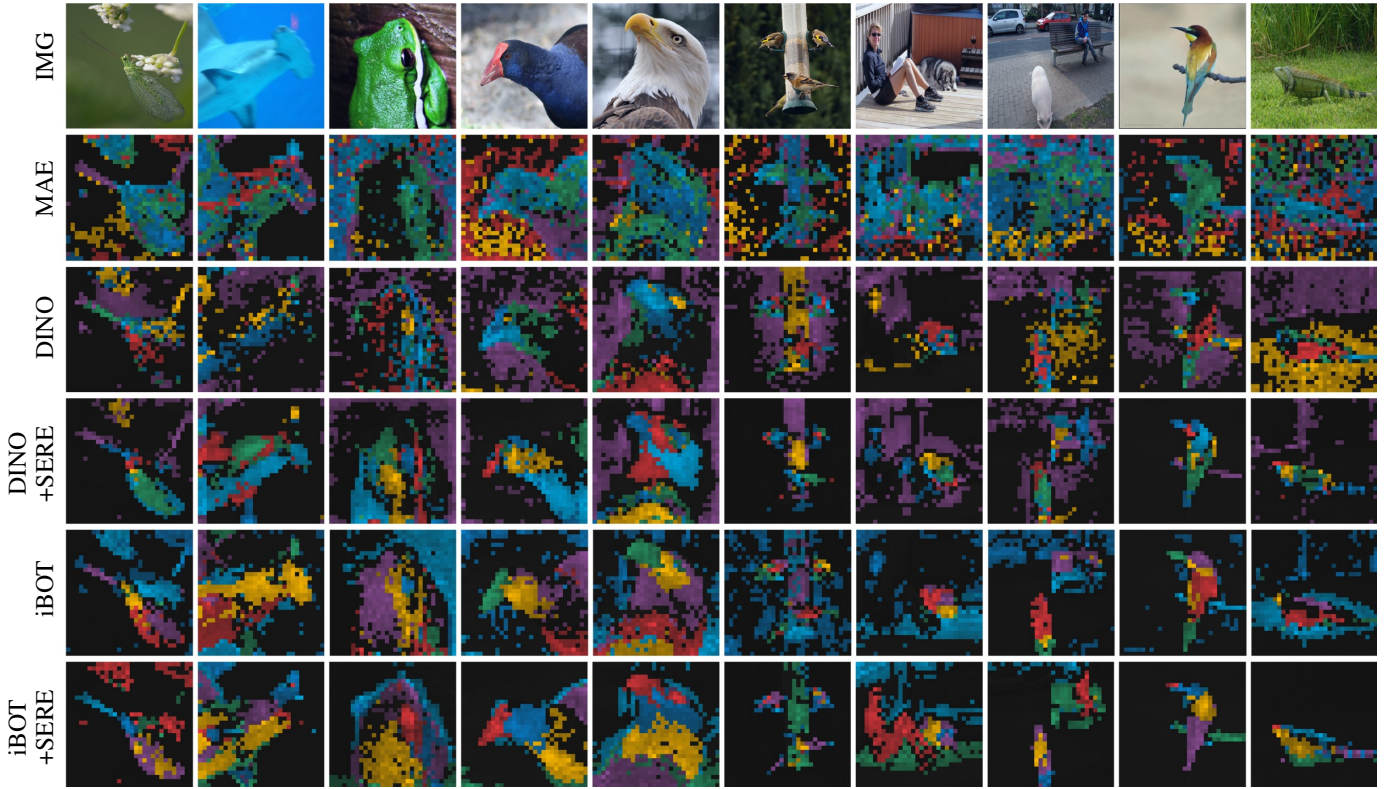
\begin{figure*}[t]
	\centering
	\begin{overpic}[width=\linewidth]{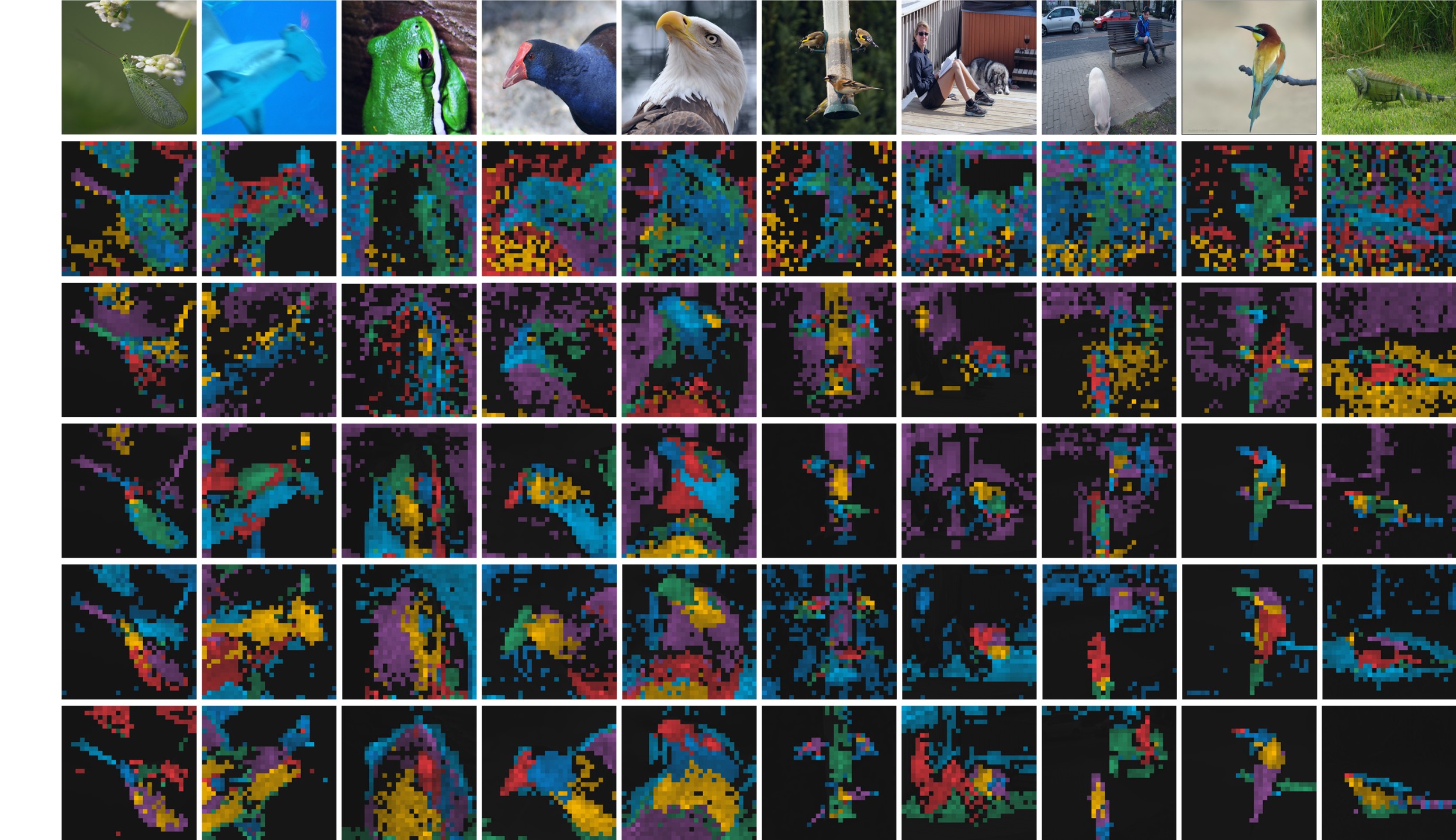}
		\put(2, 52){\rotatebox{90}{IMG}}
		\put(2, 42){\rotatebox{90}{MAE}}
		\put(2, 31.5){\rotatebox{90}{DINO}}
		\put(0, 21.5){\rotatebox{90}{DINO}}
		\put(2, 21.0){\rotatebox{90}{+SERE}}
		\put(2, 12.5){\rotatebox{90}{iBOT}}
		\put(0, 02.3){\rotatebox{90}{iBOT}}
		\put(2, 01.6){\rotatebox{90}{+SERE}}
	\end{overpic}
	\caption{Visualization for attention maps from the last block of the 
	  pre-trained ViT-S/16. 
    We extract the attention maps of the CLS token on other patch-level tokens. 
	  Different colors indicate the regions focused by different heads.
	}\label{fig:visualization_attention_maps}
\end{figure*}

\myPara{Comparison with masked image modeling~(MIM).} 
We also demonstrate that 
our proposed method, \ours, 
outperforms and complements various masked image modeling~(MIM) based methods. 
As shown in \tabref{tab:mae}, 
\ours~can significantly enhance contrastive learning based approach (e.g., DINO).
DINO+SERE achieves comparable performance compared to MIM based methods 
(iBOT and MAE), 
requiring less pre-training/fine-tuning epochs. 
Meanwhile, \ours~and MIM can be complementary. 
For instance, cooperating with \ours~further improves iBOT by 0.4\% Top-1 accuracy. 
Moreover, 
qualitative results in \figref{fig:visualization_attention_maps}  
show that \ours~produces more precise and less noisy attention maps than iBOT. 
These results strongly confirm the effectiveness of \ours~compared to MIM-based methods.

\myPara{Cooperating with more self-supervised learning methods.}
The self-relation representation is 
orthogonal to the existing feature representations.
Therefore,
it can be integrated into various self-supervised learning methods. 
To demonstrate this, we combine the \ours~with MoCo v3~\cite{Chen_2021_ICCV}, DINO, and iBOT,
\ie utilizing the self-supervision of these methods as the $L_{I}$ in \eqref{eq:loss_sum}.
We pre-train models on the ImageNet-S$_{300}$ dataset with 100 epochs to save computation costs,
and other training settings
are constant with baseline methods.
As shown in \tabref{tab:mococv3}, using \ours~consistently improves baseline methods,
verifying its generalization to different methods.
For example, \ours~improves the MoCo v3 by 1.8\% on mIoU and 2.0\% on mAcc for semantic segmentation
on the Pascal VOC dataset. 
For the semi-supervised semantic segmentation on the ImageNet-S$_{300}$ dataset, \ours~gains 5.1\% on mIoU over MoCo v3.

\begin{table}[t]
	\caption{Ablation of using different representations for self-supervised training.
	The $L_I$, $L_p$, and $L_c$ denote the loss functions using
	image-level embedding~\cite{caron2021emerging}, spatial self-relation, and channel self-relation, respectively.
	The model without these three losses is randomly initialized
	when fine-tuned on downstream tasks.}
	\centering
	\setlength{\tabcolsep}{3.5mm}
	\begin{tabular}{cccccccccccc}	\toprule
		\tRows{$L_I$}  & \tRows{$L_p$}  & \tRows{$L_c$} & \tCols{VOC SEG} & \tCols{ImageNet-S$_{300}^{\rm PT}$} \\ \cline{4-7}
		& & & mIoU            & mAcc      & val  & test    \\
		\midrule
		\xmark & \xmark & \xmark & 25.6 & 35.7 & 0.2  & 0.2 \\ \midrule
		\cmark &        &        & 68.1 & 81.1 & 28.8 & 29.6 \\
		       & \cmark &        & 71.5 & 83.0 & 23.7 & 23.7 \\
		       &        & \cmark & 61.4 & 75.6 & 22.5 & 22.3 \\ \midrule
		\cmark & \cmark &        & 70.7 & 82.6 & 33.3 & 34.5 \\
		\cmark &        & \cmark & 69.8 & 82.9 & 36.5 & 38.3 \\
					 & \cmark & \cmark & 71.5 & 83.3 & 30.6 & 30.3 \\ \midrule
		\cmark & \cmark & \cmark & \tb{73.5} & \tb{84.7} & \tb{41.2} & \tb{42.0} \\
		\bottomrule
	\end{tabular}
	\label{tab:ablation_channel_pixel}
\end{table}

\begin{table}[t]
	\caption{Segmentation F-measure~\cite{ChengPAMI} on the PASCAL VOC dataset. 
    The F-measure ignores semantic categories.}
	\centering
	\setlength{\tabcolsep}{4.0mm}
	\begin{tabular}{cccc}	
        \toprule
		& $L_p$ & $L_p+L_I$ & $L_p+L_I+L_c$ \\
		\midrule
		IoU & 87.1 & 86.7 & 87.7 \\
		\bottomrule
	\end{tabular}
	\label{tab:fmeasure}
\end{table}

\begin{table}[t]
	\caption{Cooperating self-relations with patch-level embeddings.
	DINO+ indicates adding the clustering loss using patch-level embeddings to \DINO. }
	\centering
        \setlength{\tabcolsep}{2.5mm}
        \begin{tabular}{cccccccc}	\toprule
        \tRows{DINO} & \tRows{DINO+} & \tRows{\ours}
                        & \tCols{VOC SEG}
                        & \tCols{ImageNet-S$_{300}^{\rm PT}$}                        \\ \cline{4-7}
                        & &                            & mIoU          & mAcc      & val & test \\
        \midrule
        \cmark & & & 68.1 & 81.1 & 28.8 & 29.6 \\
        & \cmark          &                            & 72.6          & 84.3  &    40.0 & 40.4 \\
                        & & \cmark                     & 73.5              & 84.7 & 41.2 & 42.0  \\
        \midrule
        & \cmark          & \cmark                     & \tb{75.0} & \tb{86.1} & \tb{44.8} & \tb{46.0} \\
        \bottomrule
        \end{tabular}
        \label{tab:p2p}
\end{table}

\begin{table}[t]
	\caption{Comparison with Barlow~\cite{barlow_Twins} that utilizes the batch-relation based loss.}
        \centering
        \setlength{\tabcolsep}{5.0mm}
        \begin{tabular}{lcccccc}	\toprule
                        & \tCols{VOC SEG}
                        & \tCols{ImageNet-S$_{300}^{\rm PT}$}                       \\ \cline{2-5}
                        & mIoU          & mAcc       & val & test \\
                        \midrule
                        Barlow~\cite{barlow_Twins}                        & 69.5          & 82.2    & 33.2 & 32.9  \\
                        \ours                     & \tb{69.8}              & \tb{82.9} & \tb{36.5} & \tb{38.3} \\
        \bottomrule
        \end{tabular}
        \label{tab:barlow}
\end{table}

\subsection{Ablation Studies}
\label{sec:Ablation}
To save computational costs for the ablation study,
we pre-train all models on the ImageNet-S$_{300}$~\cite{gao2021luss} dataset
with two global views for 100 epochs.
We evaluate models with 
semantic segmentation on the PASCAL VOC dataset 
and semi-supervised semantic segmentation on the ImageNet-S$_{300}$ dataset.
\begin{table}[t]
	\caption{The effect of different numbers of heads $M$  
 for spatial self-relation.}
	\centering
	\setlength{\tabcolsep}{6.0mm}
	\begin{tabular}{cccccccccc}	\toprule
		\tRows{$M$}
			& \tCols{VOC SEG}
			 & \tCols{ImageNet-S$_{300}^{\rm PT}$}                                             \\ \cline{2-5}
			& mIoU                       & mAcc         & val & test   \\
		\midrule
		1  & 72.4                       & 84.0            & 38.7   & 39.3     \\
		3  & 72.7                       & 84.8            & 38.9   & 39.4    \\
		6  & \tb{73.5}              & 84.7          & \tb{41.2} & \tb{42.0} \\
		12 & 73.4                       & \tb{85.1}    & 40.8   & 41.7    \\
		16 & 72.5                       & 84.3             & 39.3   & 39.8    \\
		\bottomrule
	\end{tabular}
	\label{tab:multi_head}
\end{table}

\begin{table}[t]
	\caption{The effect of different $t_p$ and $t_c$ in \eqref{eq:self_relation_pixel} and \eqref{eq:self_relation_channel}.
 }
	\centering
	\setlength{\tabcolsep}{4.5mm}
	\begin{tabular}{cccccccc}	\toprule
		\tRows{$t_p$} & \tRows{$t_c$}
						& \tCols{VOC SEG}
						& \tCols{ImageNet-S$_{300}^{\rm PT}$}                         \\ \cline{3-6}
						&                            & mIoU          & mAcc    & val & test        \\
		\midrule
		0.50          & 0.50                       & 72.0          & 84.2         & 36.7 & 36.7   \\
		\tb{0.50}          & \tb{0.10}                       & \tb{73.5} & \tb{84.7} & \tb{41.2} & \tb{42.0} \\
		0.50          & 0.01                       & 70.4          & 82.7              & 33.6  & 34.6   \\
		\midrule
		1.00          & 0.10                       & 70.2          & 83.1        & 36.7    & 38.2      \\
		\tb{0.50}          & \tb{0.10}                       & 73.5          & 84.7   & \tb{41.2} & \tb{42.0} \\
		0.10          & 0.10                       & \tb{73.7} & \tb{85.0}      & 39.9 & 40.8   \\
		\bottomrule
	\end{tabular}
	\label{tab:multi_tau}
\end{table}

\begin{table}[t]
	\caption{
		The effect of different $\alpha$ and $\beta$ in \eqref{eq:loss_sum} 
  when cooperating the \ours~with \iBOT. 
        All models are pre-trained for 100 epochs on ImageNet-1K.}
	\centering
	\setlength{\tabcolsep}{2.0mm}
	\begin{tabular}{cccc|cccc}	\toprule
		\thRows{$\alpha$} & \thRows{$\beta$} & \tCols{Classification} \vline & \multicolumn{4}{c}{Segmentation} \\ 
    & & \tCols{ImageNet-1K} \vline & \tCols{VOC} & 
		\tCols{ImageNet-S$_{\rm PT}$} \\ \cline{3-8}
		& & Top-1 & Top-5 & mIoU & mAcc & val & test \\	\Xhline{0.5pt}
		0.20 & 0.20 & 81.3 & 95.7 & 80.7 & 89.9 & 39.9 & 39.3 \\
		0.20 & 0.50 &\tb{81.5}&\tb{95.8}&\tb{81.2}&\tb{90.0}& 41.0 & 40.3 \\
		0.20 & 1.00 & 81.3 & 95.8 & 80.9 & 89.8 & \tb{41.7} & \tb{41.8} \\ \hline	
		0.10 & 0.50 & 81.3 & 95.8 & 80.9 & 89.5 & 40.7 & 40.5 \\
		0.20 & 0.50 &\tb{81.5}&\tb{95.8}&\tb{81.2}&\tb{90.0}&\tb{41.0}&\tb{40.3} \\
		0.80 & 0.50 & 81.3 & 95.8 & 80.8 & 89.7 & 40.3 & 40.1 \\ \bottomrule
	\end{tabular}
	\label{tab:alpha_beta}
\end{table}

\begin{table}[t]
	\caption{The effect of the asymmetric losses in~\eqref{eq:l_b} and \eqref{eq:l_d}.}
	\centering
	\setlength{\tabcolsep}{4.0mm}
	\begin{tabular}{lcccccccc}	\toprule
    & \tCols{VOC SEG} & \tCols{ImageNet-S$_{300}^{\rm PT}$} \\ \cline{2-5}
		& mIoU & mAcc & val & test \\ \midrule
		DNIO baseline & 68.1 & 81.1 & 28.8 & 29.6 \\
		+\ours~symmetry & 72.1 & 84.4 & 37.1 & 37.9 \\
		+\ours~asymmetric&\tb{73.5}&\tb{84.7}&\tb{41.2}&\tb{42.0} \\ \bottomrule
	\end{tabular}
	\label{tab:predciton}
\end{table}

\begin{table}[t]
	\caption{
The effect of self-relation representation on CNN. DINO and SERE are trained with the ResNet-50 network.}
	\centering
	\setlength{\tabcolsep}{4.0mm}
	\begin{tabular}{lcccccc}	\toprule
		& \tCols{VOC SEG}	& \tCols{ImageNet-S$_{300}^{\rm PT}$} \\ \cline{2-5}
		& mIoU & mAcc & val & test \\	\midrule
		DINO (ResNet-50) & 61.6 & 74.6 & 20.2 & 19.9 \\
		+\ours~(ResNet-50) & \tb{62.5} & \tb{75.0} & \tb{20.9} & \tb{20.7} \\
		\bottomrule
	\end{tabular}
	\label{tab:adaptability_cnn}
\end{table}

\myPara{Effect of spatial and channel self-relation.}
We compare the effectiveness of different
representation forms for self-supervised learning,
\ie our proposed spatial/channel self-relations and 
image-level feature embeddings used by DINO.
As shown in~\tabref{tab:ablation_channel_pixel},
the spatial self-relation improves the mIoU 
by 3.4\% and mAcc by 1.9\% on the PASCAL VOC dataset compared to the feature embedding.
These results show that training self-supervised ViT with spatial self-relation
further enhances the spatial relation modeling ability of ViT,
benefiting dense prediction tasks. 
Although inferior to the other two representation forms,
channel self-relation still improves the representation quality of ViT.
The model pre-trained with channel self-relation performs much better than the randomly initialized model 
on segmentation and classification tasks.

\myPara{Cooperating with image-level embeddings.}
We verify the orthogonality between self-relations and 
image-level embeddings, 
as shown in \tabref{tab:ablation_channel_pixel}.
When combined with the image-level feature embedding,
the spatial and channel self-relations improve the mIoU by 2.6\% and 1.7\% on the PASCAL VOC dataset. 
On the ImageNet-S$_{300}$ dataset, 
there is also an improvement of 4.5\% and 7.7\% on mIoU over feature embedding.
And cooperating three representations further boosts the performance on all tasks,
indicating that self-relations are orthogonal and complementary to  image-level feature embeddings for self-supervised learning. 

\myPara{Cooperation between $L_I$ and $L_c$.} 
\tabref{tab:ablation_channel_pixel} shows that $L_p$ alone performs better 
than $L_p+L_I$ or $L_p + L_c$ 
on the PASCAL VOC dataset. 
However, 
using $L_p+L_I+L_c$ performs better than $L_p$. 
This phenomenon is because 
utilizing image-level embedding~($L_I$) and 
channel self-relation~($L_c$) 
have their limits, 
while their cooperation can mitigate them. 
The details are as follows:
1) Regarding $L_c$, 
modeling channel self-relations requires meaningful and diverse channel features as the foundation. 
However, 
solely relying on $L_c$ 
cannot adequately optimize the channel features 
and may lead to model collapse, 
where an example is that each channel encodes the same features. 
In comparison, $L_I$ facilitates learning diverse and meaningful channel features, 
thus addressing the limitation mentioned above of $L_c$. 
2) The $L_I$ harms spatial features. 
We validate this by examining the F-measure~\cite{ChengPAMI} that ignores the semantic categories. 
\tabref{tab:fmeasure} shows a decrease in IoU when comparing $L_p+L_I$ with $L_I$, 
indicating that $L_I$ impairs spatial features. 
We assume $L_I$ makes representations less discriminable 
in the spatial dimension than $L_p$. 
However, 
by using $L_c$ simultaneously, 
we promote learning more accurate spatial features, 
mitigating the drawback caused by using $L_I$.

\myPara{Cooperating with patch-level embeddings.}
We also verify the orthogonality of self-relation representation to
patch-level embeddings in \tabref{tab:p2p}.
As a baseline,
we add a clustering loss using patch-level embeddings to DINO, denoted by DINO+.
DINO+ consistently advances DINO, 
showing the effectiveness of patch-level embedding. 
Compared to DINO+, 
the self-relation improves the mIoU by 0.9\% and 1.2\% on PASCAL VOC and ImageNet-S datasets. 
Cooperating two representations further brings constant improvements over DINO+,
\eg achieving 2.4\% and 4.8\% gains on mIoU for PASCAL VOC and ImageNet-S datasets.
These results indicate that the self-relation is complementary to patch-level embedding
for self-supervised ViT.

\myPara{Comparison between self-relation and batch-relation.}
A related work, 
Barlow~\cite{barlow_Twins}, models channel relation in the whole batch, \ie batch-relation.
In comparison, the proposed SERE computes self-relation within a single image.
To verify the advantage of self-relation over batch-relation, 
we pre-train the ViT-S/16 with the two forms of relation, respectively.
As shown in \tabref{tab:barlow}, 
compared to the batch-relation, 
the self-relation improves mIoU by 0.3\% and 3.3\% on the PASCAL VOC and ImageNet-S$_{300}$ datasets. 
These results show that self-relation
is more suitable for the training of ViT over batch-relation.

\myPara{Effect of multi-head.}
We utilize the multi-head spatial self-relation following the MHSA module in ViT.
\tabref{tab:multi_head} shows the effect of different numbers of heads $M$
in spatial self-relation.
Compared to the single-head version,
increasing $M$ to 6 brings the largest performance gain of 1.1\% on mIoU
for the PASCAL VOC dataset.
$M=12$ achieves limited extra gains, 
while $M=16$ suffers a rapid performance drop.
More heads enable diverse spatial self-relation,
but the number of channels used for calculating each self-relation is reduced.
Too many heads result in inaccurate estimation of self-relation,
hurting the representation quality.
So we default set the number of heads to 6
to balance the diversity and quality of spatial self-relation.

\myPara{Effect of sharpness.}
The temperature terms in~\eqref{eq:self_relation_pixel} and \eqref{eq:self_relation_channel}
control the sharpness of the self-relation distributions.
A small temperature sharpens the distributions, 
while a large temperature softens the distributions.
In \tabref{tab:multi_tau},
we verify the effectiveness of temperatures for both spatial and channel self-relations.
For the channel self-relation,
decreasing temperature from 0.1 to 0.01
results in a rapid performance drop from 73.5\% to 70.4\% on mIoU
for the PASCAL VOC dataset.
And increasing it from 0.1 to 0.5 also degrades the mIoU from 73.5\% to 72.0\%.
Therefore, we choose 0.1 as the default temperature for the channel self-relation.
For the spatial self-relation,
the temperature 0.5 performs better than 1.0,
and changing the temperature from 0.5 to 0.1 has a limited difference.
We set the default temperature of spatial self-relation to 0.5 because
a temperature of 0.5 achieves slightly better performance on
the large-scale ImageNet-S dataset. 

\myPara{Effect of loss weights.}
The $\alpha$ and $\beta$ in Equ. (6) 
determine the relative importance of spatial and channel self-relations, respectively. 
\tabref{tab:alpha_beta} shows that the \ours~is robust to different $\alpha$ and $\beta$. 
Among different weights, 
the combination of $\alpha = 0.2$ and $\beta=0.5$ 
achieves the best performance on the classification task and 
competitive performances on the segmentation task. 
Therefore, we use this combination as the default setting.

\myPara{Effect of asymmetric loss.}
The asymmetric structure has been proven effective for non-contrastive loss~\cite{byol,Chen_2021_CVPR} 
when using image-level embedding as the representation.
To verify if self-relation representations also
benefit from the asymmetric structure,
we compare 
the asymmetric and symmetry structures for the self-relation based loss in \tabref{tab:predciton}.
Self-relation improves the DINO baseline
with both asymmetric and symmetry structures.
The symmetrical structure outperforms the DINO on PASCAL VOC and ImageNet-S$_{300}$ datasets with 4.0\% and 8.3\% on mIoU.
The asymmetric structure further advances symmetric structure 
by 1.4\% and 4.1\% on mIoU for the PASCAL VOC and ImageNet-S$_{300}$ datasets.
Therefore, though the asymmetric structure is not indispensable
for self-relation,
it still benefits the pre-training with self-relation.

\myPara{Adaptability to convolutional neural networks.}
Using self-relation for self-supervised learning is inspired by 
the properties of ViT.
Still, we wonder if the self-relation representation could
benefit self-supervised learning 
on convolutional neural networks (CNN).
To verify this, 
we pre-train the ResNet-50~\cite{he2016deep} with DINO and SERE, respectively.
The training details are shown in the supplementary.
As shown in \tabref{tab:adaptability_cnn}, 
SERE improves DINO by 0.7\% and 0.8\% on mIoU for
the semantic segmentation task on the PASCAL VOC and ImageNet-S$_{300}$ datasets compared to DINO. 
Though designed for ViT, the self-relation still improves the representation quality of the CNN. 
Meanwhile, the improvement on CNN is relatively small compared to that on ViT, 
showing that the self-relation is more suitable for ViT.

\begin{figure}[t]
	\centering
	\begin{overpic}[width=1.0\linewidth]{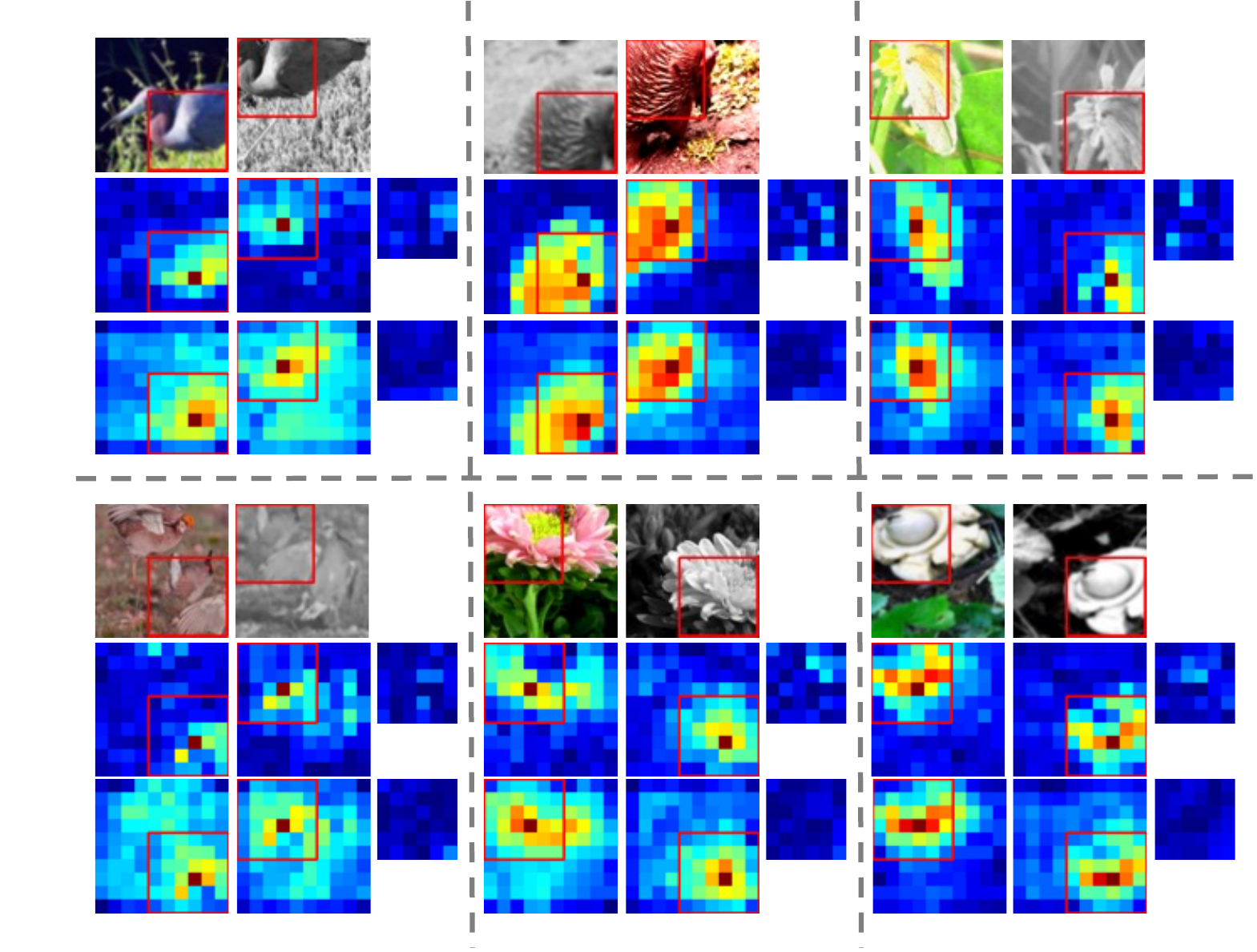}
		\put(0, 29.5){(a)}
		\put(0, 18.5){(b)}
		\put(0, 7.7){(c)}

		\put(0, 66){(a)}
		\put(0, 55.0){(b)}
		\put(0, 44.0){(c)}

		\put(8.5, -3.5){view1}
		\put(19.5, -3.5){view2}
		\put(31.5, -3.5){$\Delta$}

		\put(39.7, -3.5){view1}
		\put(50.5, -3.5){view2}
		\put(63.0, -3.5){$\Delta$}

		\put(70.5, -3.5){view1}
		\put(81.5, -3.5){view2}
		\put(93.5, -3.5){$\Delta$}

	\end{overpic}
	\vspace{0pt}
	\caption{The differences between spatial self-relations of two views.
		(a) Two views from each image.
		(b) The spatial self-relation generated by DINO.
		(c) The spatial self-relation generated by \ours.
		View1 and view2 mean the self-relations of two views generated from an image.
		The $\Delta$ is the difference between self-relations in the overlapping region, which is indicated by red boxes.
		We give the details of the visualization method in the supplementary.
	}\label{fig:vis_diff}
\end{figure}

\begin{figure}[t]
	\centering
        \hspace{3pt}
	\begin{overpic}[width=0.98\linewidth]{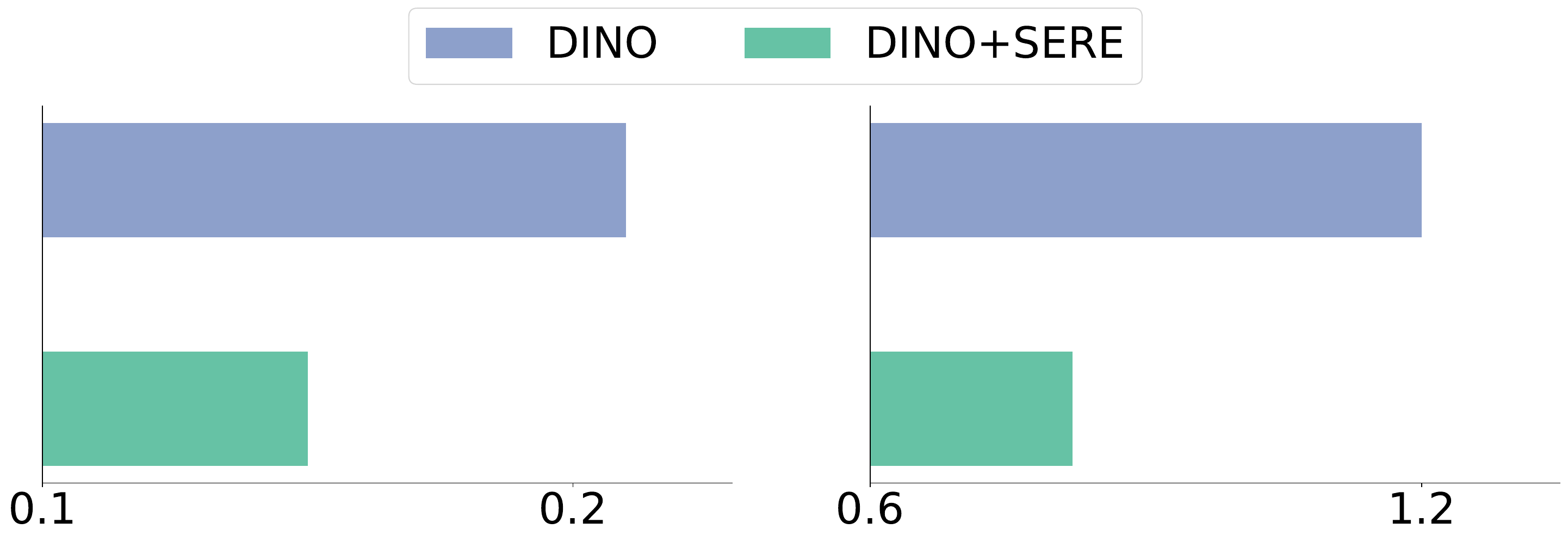}
	\put(40.8, 22.0){\footnotesize 0.21}
	\put(20.5, 7.3){\footnotesize 0.15}

	\put(91.2, 22.0){\footnotesize 1.20}
	\put(69.0, 7.1){\footnotesize 0.82}

	\put(-0.9, 11.0){\rotatebox{90}{\small Spatial}}
	\put(52.3, 10.0){\rotatebox{90}{\small Channel}}
	\put(15.5, -2){\small Difference}
	\put(68.0, -2){\small Difference}
	\end{overpic}
	\caption{The average differences of spatial~(left) and channel~(right) 
	  self-relations between two views on the val set of ImageNet-S. 
	  We show the calculation details in the supplementary.
	}\label{fig:anaysis_diff}
\end{figure}
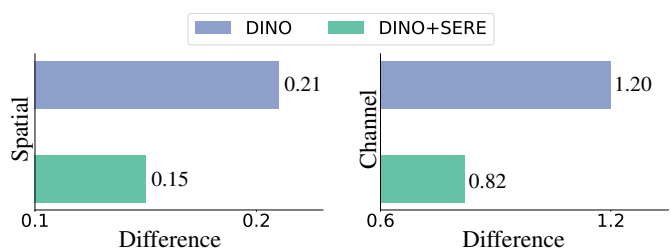

\subsection{Analysis and Visualization}
\myPara{Invariance on self-relations.}
The importance of learning representations invariant to image augmentations, 
\eg 
scaling, shifting, and color jitter, 
has been validated in self-supervised 
learning~\cite{purushwalkam2020demystifying,Patrick_2021_ICCV,Misra_2020_CVPR,bardes2022vicreg,ericsson2021selfsupervised,Wang_2017_ICCV}. 
However, 
existing methods focus 
on the invariance of feature embeddings 
but do not 
consider the invariance of spatial/channel relations, 
which are also important properties of ViT. 
In contrast, 
our proposed \ours~can enhance 
the invariance of spatial/channel relations.
To verify this,
we measure the averaged differences between self-relations of different views. 
As shown in \figref{fig:anaysis_diff}, 
we obverse that \ours~significantly
narrows the self-relation differences 
in both the spatial and channel dimensions. 
The visualizations in \figref{fig:vis_diff}
also show that 
the SERE pre-trained model produces 
smaller spatial self-relation differences
on the overlapping regions of two views. 
A smaller difference means a higher invariance. 
Thus, these results indicate that \ours~makes the ViT 
capture self-relations with stronger invariance to image augmentations.

\myPara{Visualization of attention maps.}
In \figref{fig:visualization_attention_maps}, 
we visualize the attention maps 
from the last block of ViT. 
These visualizations demonstrate that 
\ours~produces more precise and less noisy attention maps than 
various methods, including MIM-based methods, \ie \MAE and \iBOT. 
MAE produces noisy attention maps that highlight almost all tokens in an image. 
In comparison, 
the attention maps of \ours~mainly focus on semantic objects. 
For instance, 
the third column of \figref{fig:visualization_attention_maps} shows
that \ours~can locate the frog, but 
MAE primarily focuses on the background. 
Moreover, 
compared to iBOT and DINO, 
\ours~generates attention maps 
that locate objects more accurately. 
For instance, 
in the seventh and eighth columns of \figref{fig:visualization_attention_maps}, 
\ours~discovers the persons missed by iBOT.

\myPara{Comparison between spatial self-relation and MIM.} 
Both spatial self-relation and MIM act on the spatial dimension, 
but their effects significantly differ. 
MIM enhances the {token-level representations}, 
while spatial self-relation focuses on improving the ability to model {inter-token relations}.
We support this argument with the following points: 
1) As depicted in \figref{fig:visualization_attention_maps}, 
\ours~generates  more precise and less noisy attention maps than 
\MAE and \iBOT. 
The attention maps of ViT can reflect the ability to model inter-token relations 
because attentions are calculated as token-level relations between query and key. 
Thus this observation indicates that 
\ours~provides models with a stronger ability to capture inter-token relations. 
In \figref{fig:anaysis_diff}, we show that 
\ours~enhances the invariance of spatial self-relation 
to different image augmentations.  
3) As shown in \tabref{tab:mae}, SERE achieves consistent improvements compared to different MIM-based methods, 
strongly confirming the effectiveness of \ours~compared to MIM. 
For example, 
cooperating with \ours~improves iBOT by 0.4\% Top-1 accuracy, 
as shown in \tabref{tab:com_imagenets}.

\section{Conclusions}
In this paper, we propose a feature self-relation based self-supervised learning scheme to enhance the relation modeling ability of self-supervised ViT.
Specifically, instead of directly using feature embedding as the representation, we propose to use spatial and channel self-relations of features as representations for self-supervised learning.
Self-relation is orthogonal to feature embedding and further boosts existing self-supervised methods.
We show that feature self-relation improves the self-supervised ViT
at a fine-grained level, 
benefiting multiple downstream tasks, 
including image classification, semantic segmentation, object detection, and instance segmentation.

\myPara{Acknowledgements.} 
This work is funded by NSFC (NO. 62225604, 62176130), 
and the Fundamental Research Funds for the Central Universities 
(Nankai University, 070-63233089). 
Computation is supported by the Supercomputing Center of Nankai University.

\bibliographystyle{IEEEtran}
\bibliography{egbib}

\newcommand{\AddPhoto}[1]{{\includegraphics[width=1in,keepaspectratio]{figure/Authors/#1}}}
\newcommand{\AuthorBio}[3]{\vspace{-.2in}\begin{IEEEbiography}[\AddPhoto{#1}]{#2}#3\end{IEEEbiography}}

\AuthorBio{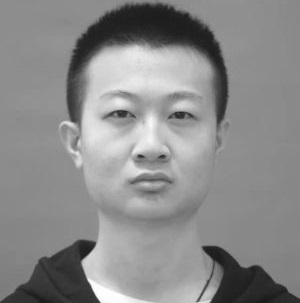}{Zhong-Yu Li}{
	is a Ph.D. student from the college of computer science, Nankai university.
	He is supervised via Prof. Ming-Ming cheng.
	His research interests include deep learning, machine learning and computer vision.
}

\AuthorBio{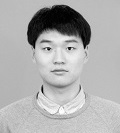}{Shanghua Gao}{
	is a Ph.D. candidate in Media Computing Lab at Nankai University.
	He is supervised via Prof. Ming-Ming Cheng.
	His research interests include computer vision and representation learning.
}

\AuthorBio{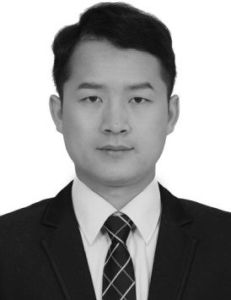}{Ming-Ming Cheng}{
	received his PhD degree from Tsinghua University in 2012,
	and then worked with Prof. Philip Torr in Oxford for 2 years.
	Since 2016, he is a full professor at Nankai University, leading the
	Media Computing Lab.
	His research interests include computer vision and computer graphics.
	He received awards, including ACM China Rising Star Award,
	IBM Global SUR Award, \etc.
	He is a senior member of the IEEE and on the editorial boards of
	IEEE TPAMI and IEEE TIP.
}

\vfill

\clearpage
\end{document}